# Retrieve-Refine-Calibrate: A Framework for Complex Claim Fact-Checking

Mingwei Sun [1, &], Qianlong Wang [2, &], and Ruifeng Xu [1, a,*]

[1] Harbin Institute of Technology (Shenzhen), Shenzhen, China;

[2] Shenzhen Technology University, Shenzhen, China;

[a] xuruifeng@hit.edu.cn

[&] These authors contributed equally to this work

**Abstract.** Fact-checking aims to verify the truthfulness of a claim based on the retrieved evidence. Existing methods typically follow a decomposition paradigm, in which a claim is broken down into sub-claims that are individually verified. However, the decomposition paradigm may introduce noise to the verification process due to irrelevant entities or evidence, ultimately degrading verification accuracy. To address this problem, we propose a *Retrieve-Refine-Calibrate* (RRC) framework based on large language models (LLMs). Specifically, the framework first identifies the entities mentioned in the claim and retrieves evidence relevant to them. Then, it refines the retrieved evidence based on the claim to reduce irrelevant information. Finally, it calibrates the verification process by re-evaluating low-confidence predictions. Experiments on two popular fact-checking datasets (HOVER and FEVEROUS-S) demonstrate that our framework achieves superior performance compared with competitive baselines.

## 1. Introduction

Fact-checking is a task of assessing the veracity of a claim based on retrieved evidence[1]. Unlike information retrieval or document classification, fact-checking requires fine-grained reasoning based on claims and evidence[2]. With misleading information on social media platforms becoming increasingly widespread[3, 4], automatic fact-checking has received considerable attention[6].

Traditional methods fine-tune specialized language models to jointly encode claims and evidence for truthfulness prediction[7, 14]. However, they suffer from restricted reasoning capacity due to insufficient linguistic comprehension[8], especially for complex claims. Recently, large language model (LLM)-based reasoning methods have become increasingly popular. These methods decompose claims[9, 10] or ask intermediate questions[11, 20], thereby enabling multi-hop reasoning. Their progress benefits from the strong semantic understanding capabilities of LLMs[24].

Although LLM-based reasoning methods have achieved notable progress in fact-checking, they still suffer from two limitations. First, claim decomposition does not always yield benefits. Instead, it may introduce noise[26], ultimately degrading fact-checking performance. For example, in Fig. 1, the decomposition mistakenly links the actor in *"East/West"* to an unrelated individual *"John Hanson"*, thus adding noise to the reasoning process. Second, the retrieved evidence usually contains irrelevant information. As shown in Fig. 1, the framework retrieves a comedy film featuring *"Anatoly Ravikovich"*. However, this film is not the one referred to in the original claim, which ultimately misleads the verification. As a result, decomposition-based approaches may produce incorrect predictions, especially for complex claims involving multiple entities or interconnected multi-hop evidence.

Rather than eliminating multi-step reasoning, our framework reconsiders how evidence is constructed prior to verification. We argue that the limitations of decomposition-based approaches stem not from insufficient reasoning capability, but from noisy and error-prone intermediate representations introduced by explicit claim splitting. To avoid these limitations, we propose the *Retrieve-Refine-Calibrate* (RRC) framework, which avoids claim decomposition and performs fact-checking through evidence retrieval, refinement, and verification calibration. Specifically, our framework contains three modules: (1) *entity-based retrieval* module, which identifies key entities in the claim and retrieves relevant evidence; (2) *claim-specific evidence refinement* module, which aims to remove redundancy and irrelevant evidence, thus reducing noise that could affect verification[12]; (3) *confidence-based calibration* module, which first verifies the claim using the refined evidence, then performs a calibration based on previous prediction confidence. Such calibration can effectively mitigate hallucinations and enhance reliability[28]. By adopting a claim-specific perspective to integrate entity-based evidence, RRC mitigates noise introduced by claim decomposition and effectively filters out irrelevant or redundant information, thereby improving verification accuracy.

Our main contributions are as follows:

(1) We revisit LLM-based fact-checking from a *decomposition-free* perspective and show that directly reasoning over claim-specific evidence can effectively avoid noise introduced by claim decomposition.

(2) We propose the *Retrieve–Refine–Calibrate* framework, which integrates entity-based retrieval, claim-specific evidence refinement, and confidence-based calibration to improve robustness and reliability in fact-checking.

(3) Extensive experiments on HOVER and FEVEROUS-S demonstrate that our framework consistently outperforms recent decomposition-based fact-checking methods.

| Claim |
|---|
| The actor who starred in East/West was also a director. He costarred with Anatoly Ravikovich in a comedy film. **(Supports)** |
| **Verification Process** |
| **Intermediate Questions:**<br>• Who is the actor who starred in *East/West*? → Retrieved: "*John Hanson*". (incorrect entity, introduces decomposition noise)<br>• What is the comedy film in which *John Hanson* costarred with *Anatoly Ravikovich*? → Retrieved: No comedy film. (irrelevant evidence, introduces retrieval noise)<br>**Rewriting:**<br>• *John Hanson*, who starred in *East/West*, was also a director. *John Hanson* costarred with *Anatoly Ravikovich* in a comedy film.<br>**Decomposed Sub-claim:**<br>• *John Hanson* was a director. → *True*<br>• *John Hanson* costarred with *Anatoly Ravikovich* in a comedy film. → *False*<br>**Final Verdict: Refutes** ✗ |

Fig. 1 An example illustrating verification errors (*i.e.*, red font) due to decomposition and irrelevant evidence retrieval in LLM-based reasoning method[20].

## 2. METHODOLOGY

### 2.1 Task Definition

Automatic fact-checking aims to verify the veracity of a claim based on the retrieved evidence. Given a claim $C$ and the evidence source $S$, a fact-checking model $M$ aims to determine a result $Y \in \{supports, refutes\}$. Here, *supports* denotes that the claim can be confirmed as true by sufficient evidence, and *refutes* indicates that the claim is contradicted or cannot be verified due to insufficient evidence.

### 2.2 Entity-based Retrieval

Understanding the evidence associated with entities in a claim is crucial for accurate fact-checking[13] since entity-level misinterpretations may result in erroneous conclusions. To this end, we adopt LUKE[27] to identify a set of entities $E=\{e_1, e_2,…,e_n\}$ from the claim, such as named persons, locations, and organizations. For each entity $e_i \in E$, we use BM25[23] to retrieve a corresponding set of evidence $E=\{E_1,E_2,…,E_n\}$ from an external knowledge source $S$.

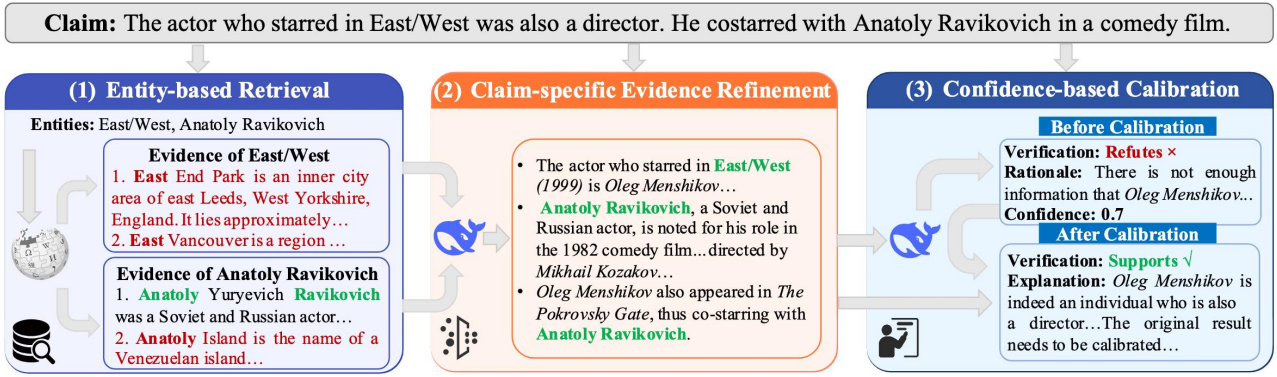

Fig. 2 Overview of our framework RRC. The framework consists of three main modules: (1) *Entity-based Retrieval* module, (2) *Claim-specific Evidence Refinement* module, and (3) *Confidence-based Calibration* module.

## 2.3 Claim-specific Evidence Refinement

While retrieved evidence provides background knowledge necessary for fact-checking, it often includes redundant or irrelevant content. For example, as illustrated in Fig. 2, the retrieved results for *"Anatoly Ravikovich"* contain unrelated entities sharing the same name, such as other *"geographic"* locations. To this end, we introduce an evidence refinement module powered by an LLM. Given the original claim C and the retrieved evidence E, the module first filters out irrelevant evidence. Then, instead of simply retaining the remaining passages, it further transforms them by compressing and paraphrasing into a concise, claim-relevant paragraph $K^*$.

## 2.4 Confidence-based Calibration

Although evidence refinement effectively reduces noise in the retrieved evidence, the inherent hallucinations of large language models may still result in erroneous verification outcomes, especially when evidence is incomplete or ambiguous. To alleviate this problem, we introduce a calibration module that uses the same LLM to re-evaluate the initial prediction, thereby improving the reliability and robustness of fact-checking results. This module comprises two components:

*Verifier ($\mathcal{M}_v$)* is an LLM employed to assess the truthfulness of a given claim based on the refined $K^*$. Specifically, we design a fact-checking prompt that instructs $\mathcal{M}_v$ to produce three outputs simultaneously: *the predicted label $Y_0$, the reasoning process R*, and *a confidence score $\gamma \in [0,1]$*:

$$(Y_0, R, \gamma) = \mathcal{M}_v(C, K^*) \tag{1}$$

Previous studies[28, 29] have shown that low-confidence predictions from large language models are often correlated with hallucinated reasoning. Motivated by this observation, we leverage prediction confidence as a signal to selectively trigger calibration, rather than applying correction to all instances. In this setting, a low confidence score indicates potential flaws in the model's reasoning, which are often caused by insufficient, ambiguous, or noisy evidence, as illustrated in Fig. 2.

*Calibrator ($\mathcal{M}_c$)* is responsible for inspecting the original reasoning R when the confidence score $\gamma$ falls below a predefined threshold $\theta$. It identifies potential logical inconsistencies or unsupported conclusions and reassesses the veracity of the claim based on the claim C, the initial prediction $Y_0$, the reasoning R, and the refined evidence $K^*$:

$$Y^* = \begin{cases} \mathcal{M}_c(C, Y_0, R, K^*), & \text{if } \gamma < \theta, \\ Y_0, & \text{otherwise}. \end{cases} \tag{2}$$

When the model exhibits hallucinations or engages in unreliable reasoning, its output is usually associated with low confidence[5]. As a reasoning-correction module, the calibrator $\mathcal{M}_c$ mitigates such uncertainty by revising flawed reasoning only when necessary, thereby reducing unreliable predictions while preserving stable and confident verification results.

## 3. EXPERIMENTS

### 3.1 Experiment Settings

*Dataset.* Experiments are conducted on two popular benchmarks: HOVER[21] and FEVEROUS-S[9]. HOVER focuses on claims that require multi-hop reasoning, including 2-hop, 3-hop, and 4-hop claims. FEVEROUS-S is a simplified subset of the FEVEROUS benchmark, consisting of simplified claims that span multiple claim types, including entity-centric and relational claims, and emphasizing evidence grounding and verification robustness. Both datasets are widely used for fact-checking tasks.

Table 1. Main experimental results (Macro-F1 score, \%). The best score is marked in **bold**, while the second-best score is marked with underline. The symbol # is from [9], † denotes our reproduction with DeepSeek-Coder for fair comparison.

| Method | HOVER | | | FEVEROUS-S |
| --- | --- | --- | --- | --- |
| | 2-hop | 3-hop | 4-hop | |
| *Pretrained methods* | | | | |
| BERT-FC[#][7] | 50.68 | 49.86 | 48.57 | 51.67 |
| LisT5[#][14] | 52.56 | 51.89 | 50.46 | 54.15 |
| *Finetuned methods* | | | | |
| RoBERTa-NLI[#][15] | 63.62 | 53.99 | 52.40 | 57.80 |
| DeBERTaV3-NLI[#][16] | 68.72 | 60.76 | 56.00 | 58.81 |
| MULTIVERS[#][17] | 60.17 | 52.55 | 51.86 | 56.61 |
| *LLM-based ICL methods* | | | | |
| Codex[#][18] | 65.07 | 56.63 | 57.27 | 62.58 |
| FLAN-T5[#][19] | 69.02 | 60.23 | 55.42 | 63.73 |
| *LLM-based reasoning methods* | | | | |
| ProgramFC[†][9] | <u>69.78</u> | 60.27 | 56.41 | 65.59 |
| FOLK[†][10] | **71.82** | 61.30 | 54.89 | 64.07 |
| Bootstrapping[†][11] | 68.45 | 61.29 | <u>59.23</u> | 66.12 |
| BiDev[†][20] | 67.85 | <u>62.03</u> | 58.87 | <u>68.61</u> |
| **RRC (Ours)** | 69.33 | **63.98** | **60.95** | **72.55** |

*Baselines.* We compare our framework with eleven baselines, categorized into four groups: 1) Pretrained methods contain BERT-FC[7], LisT5[14], , which fine-tune pretrained language models to jointly encode claims and evidence for verification. 2) Finetuned methods include RoBERTa-NLI[15], DeBERTaV3-NLI[16], MULTIVERS[17], which leverage large-scale natural language inference or document-level supervision to improve claim verification performance. 3) LLM-based in-context learning (ICL) methods include Codex[18] and FLAN-T5[19], which perform fact-checking via prompting without task-specific finetuning. 4) LLM-based reasoning methods include ProgramFC[9], FOLK[10], Bootstrapping[11], BiDeV[20], , which explicitly introduce intermediate reasoning steps, such as claim decomposition, program-guided verification, or iterative reasoning refinement, to handle complex multi-hop claims.

*Implementation Details.* All experiments are conducted in an open-book setting to simulate real-world fact-checking scenarios. We adopt DeepSeek-Coder[22] as the backbone language model for our framework. For document retrieval, we employ BM25[23] to retrieve the *top−k* most relevant documents from Wikipedia provided by each benchmark. The $k$ is set to 5 for FEVEROUS-S and 10 for HOVER. For evidence retrieval, we extract up to three entities from each claim. The threshold $\theta$ is set to 0.90 for HOVER and 0.85 for FEVEROUS-S. For a fair comparison, we apply the same retrieval configuration and the same base model for all LLM-based reasoning baselines. Following the previous work[9, 11], we use Macro-F1 as the main evaluation metric.

## 3.2 Main Results

Table 1 presents the main results. Our framework demonstrates superior performance in fact-checking complex claims. For example, compared with Bootstrapping[11], it improves Macro-F1 by 2.93% on average, mainly attributable to global claim analysis that mitigates error propagation in the Bootstrapping framework. It also surpasses BiDeV[20] by 2.36% on average. This shows that our framework, independent of claim decomposition, effectively mitigates the potential noise that such a process may introduce. Besides, the improvement on the 2-hop subset of HOVER is limited in our framework, mainly because such claims decompose more easily, favoring methods based on the decomposition paradigm. Nevertheless, our framework exceeds the suboptimal model by 1.95%@3-hop and 1.72%@4-hop of HOVER, further demonstrating its potential for verifying complex claims.

Table 2. Ablation study. *w/o Entity Identifier* uses full-claim retrieval; *w/o Evidence Refinement* uses unrefined evidence; *w/o Calibrator* uses the initial prediction without calibrating.

| Method | HOVER | | | FEVEROUS-S |
|---|---|---|---|---|
| | 2-hop | 3-hop | 4-hop | |
| RRC | **69.33** | **63.98** | **60.95** | **72.55** |
| *w/o Entity Identifier* | 69.15 | 63.37 | 60.34 | 63.46 |
| *w/o Evidence Refinement* | 66.15 | 60.29 | 57.50 | 70.41 |
| *w/o Calibrator* | 65.19 | 61.69 | 60.53 | 69.57 |

## 3.3 Ablation Study

Table 2 shows the ablation results. We can see that removing the entity identifier leads to a noticeable performance drop, particularly on FEVEROUS-S, highlighting its critical role in guiding accurate evidence retrieval. Excluding the evidence refinement module significantly degrades performance on HOVER 3-hop and 4-hop subsets, suggesting that refining retrieved evidence is especially important for handling complex claims that require multi-hop reasoning. In addition, removing the calibrator module reduces performance, which demonstrates its effectiveness in mitigating hallucinations and stabilizing verification results in fact-checking.

| **Claim:** Quinceañera inspired the TV show that the band was formed on. Jack Duarte was a member of this band. | **Supports** |
|---|---|
| **The strongest baseline BiDev:**<br>**Intermediate Questions:**<br>• What is the TV show that the band was formed on? → *"Coca-Cola Popstars"*.<br>• What is the name of the band that Jack Duarte was a member of ? → *"Eme 15"*.<br>**Rewriting:**<br>• Quinceañera inspired the TV show *"Coca-Cola Popstars"* that the band *"Eme 15"* was formed on. Jack Duarte was a member of *"Eme 15"*.<br>**Decomposed Sub-claim:**<br>• The event Quinceañera inspired the TV show *"Coca-Cola Popstars"*. → True<br>• Jack Duarte was a member of *"Eme 15"*. → True<br>• The band *Eme 15* was formed on the TV show *"Coca-Cola Popstars"*. → False<br>**Final Verdict: Refutes** ✗ | **Our framework RRC:**<br>**Entity-based Retrieval:**<br>• Quinceañera: Quinceañera is a Mexican telenovela produced by Carla Estrada for Televisa in 1987…<br>• Jack Duarte: Robert Jack Duarte Wallace (born April 7, 1986 in Mexico City, Distrito Federal) is a Mexican actor and singer… *Duarte* High School is a public high school located in *Duarte*.<br>**Claim-specific Evidence Refinement:**<br>• The TV series Miss XV was explicitly inspired by the 1987 telenovela Quinceañera…<br>Jack Duarte was a member of Eme 15.<br>**Confidence-based Calibration:**<br>• Verification: Supports    Confidence: 0.9<br>• Rationale: Miss XV was inspired by the telenovela Quinceañera…The refined evidence confirms both key components of the claim.<br>• Calibration reason: No inconsistencies or reasoning flaws are present in the model's assessment.<br>**Final Verdict: Supports** ✓ |

Fig. 3 Case study. Red indicates errors introduced by decomposition (left box) or noise in retrieved evidence (right box).

## 3.4 Analysis and Discussion

*Analysis of Backbone Models.* Table 3 presents a comparison between our method and a representative state-of-the-art approach under a fixed backbone model, Qwen3-32B, which allows us to isolate the effect of the framework design. As shown, RRC consistently outperforms BiDev across all settings, with more pronounced gains on complex claims, particularly on the 4-hop subset of HOVER and on FEVEROUS-S. Moreover, the performance improvements of RRC remain consistent across backbone models of different sizes, indicating that our proposed framework generalizes well beyond a specific model scale. Removing the calibrator module also leads to a clear performance drop, further confirming the effectiveness of confidence-based calibration in stabilizing verification performance. Overall, the results demonstrate that the performance gains of RRC arise primarily from the proposed framework rather than the choice or size of the backbone model itself.

Table 3. Comparison with state-of-the-art method using Qwen3-32B as the backbone model.

| Method | HOVER | | | FEVEROUS-S |
| --- | --- | --- | --- | --- |
| | 2-hop | 3-hop | 4-hop | |
| BiDev | 64.41 | 60.74 | 56.58 | 66.92 |
| **RRC (Ours)** | **65.77** | **61.24** | **60.48** | **69.83** |
| RRC *w/o Calibration* | 62.34 | 59.25 | 56.71 | 65.48 |

*Analysis of Hyperparameters.* We analyze the impact of three hyperparameters: the number of entities for retrieval $n$, the number of retrieved documents $k$, and the confidence threshold $\theta$. The analysis results are shown in Fig. 4. On HOVER, the overall performance remains relatively stable as $n$ varies, indicating that the framework is not overly sensitive to the exact number of extracted entities. In contrast, increasing $k$ consistently improves performance up to a certain point, as retrieving more documents enhances evidence coverage and reduces the risk of missing critical supporting facts. With respect to the confidence threshold, performance peaks at $\theta = 0.90$, reflecting an effective trade-off between correcting unreliable predictions and avoiding unnecessary recalibration. On FEVEROUS-S, a different trend is observed. Performance initially increases and then declines as $n$, $k$, and $\theta$ increase. This indicates that excessive entity extraction or document retrieval may introduce redundant or noisy evidence, which can negatively affect verification accuracy. When $n=3$, $k=5$, $\theta=0.85$, evidence coverage and noise mitigation are best balanced.

Table 4. Quality evaluation of original evidence and refined version. *w/o ER (Evidence Refinement module)* denotes the original retrieved evidence, and *w/ ER* is the refined version.

| Evaluation Dimension | HOVER | | FEVEROUS-S | |
| --- | --- | --- | --- | --- |
| | *w ER* | *w/o ER* | *w ER* | *w/o ER* |
| Factuality | 2.66 | 2.66 | **3.25** | 2.89 |
| Fluency | **4.12** | **4.12** | 4.04 | **4.49** |
| Conciseness | 2.32 | 2.32 | **2.59** | 2.48 |
| Completeness | 2.38 | 2.38 | **3.19** | 2.58 |
| Average | **3.45** | 2.87 | **3.27** | 3.43 |

*Analysis of Refined Evidence Quality.* To evaluate the quality of the refined evidence, we conduct assessments on both the original evidence and the refined version using the golden evidence as a reference. Following[25], we employ GPT-4o as the evaluator and measure four dimensions: *Factuality, Fluency, Conciseness*, and *Completeness*. As shown in Table 4, refined evidence achieves notable improvements in factuality, conciseness, and completeness, indicating the effectiveness of the claim-specific evidence refinement module. However, in terms of fluency, the refined evidence lags behind the original retrieval on both datasets. This may result from the refinement process emphasizing factual consistency and completeness, which can inadvertently reduce linguistic naturalness and fluency.

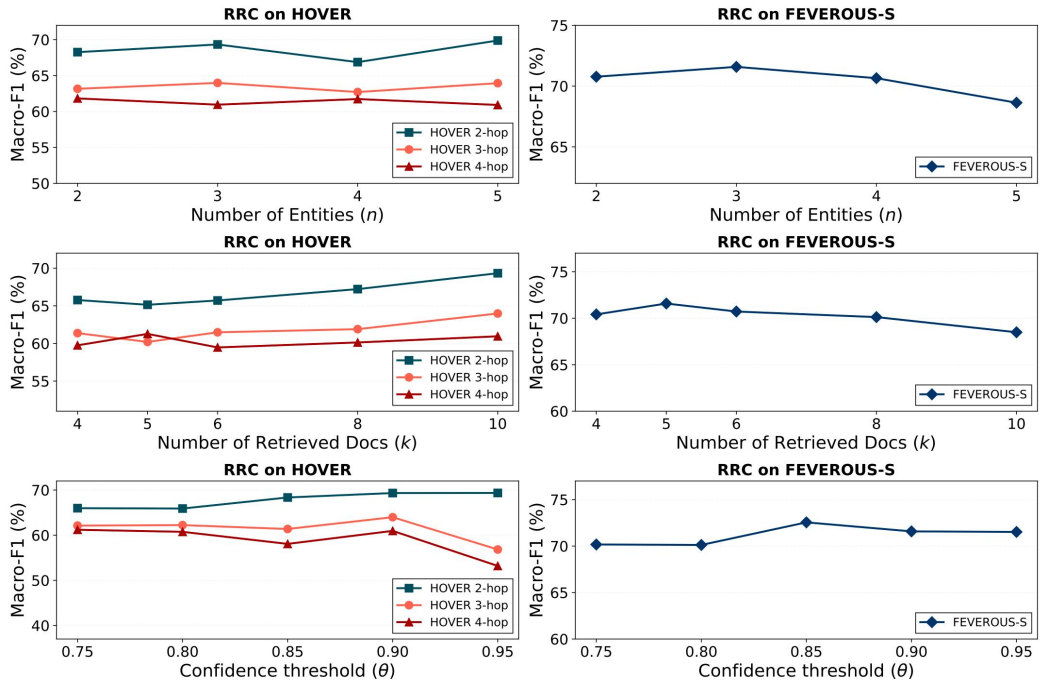

Fig. 4 Analysis across different hyperparameters.

*Case Study.* To illustrate the advantages of our framework, we present a case study. As shown in Fig. 3, our framework correctly verifies the claim using *entity-based retrieval* and *claim-specific evidence refinement* module. In contrast, although the strongest baseline BiDeV retrieves the correct band for *"Jack Duarte"*, the decomposition process incorrectly introduces an irrelevant entity *"Coca-Cola Popstars",* which ultimately leads to an incorrect prediction. This example highlights that the decomposition paradigm is vulnerable to noise introduced during intermediate reasoning, especially when handling complex claims involving multiple entities. By avoiding explicit decomposition and focusing on claim-level evidence integration, our framework demonstrates improved accuracy and robustness in such complex fact-checking scenarios.

## 4. Summary

We propose an effective framework for verifying complex claims based on LLM, named RRC. The RRC alleviates the limitations of the claim decomposition paradigm-based method by incorporating entity-based retrieval, claim-specific evidence refinement, and confidence-based calibration. Extensive experiments on HOVER and FEVEROUS-S show superior performance over baselines, highlighting the effectiveness of our framework for accurate and robust fact verification and offering a promising direction for accurate fact-checking.